# Built Infrastructure Monitoring and Inspection Using UAVs and Vision-based Algorithms

**Khai Ky Ly and Manh Duong Phung**
Western Sydney University
255 Elizabeth Street, Sydney, NSW 2000

**Abstract**
This study presents an inspecting system using real-time control unmanned aerial vehicles (UAVs) to investigate structural surfaces. The system operates under favourable weather conditions to inspect a target structure, which is the Wentworth light rail base structure in this study. The system includes a drone, a GoPro HERO4 camera, a controller and a mobile phone. The drone takes off the ground manually in the testing field to collect the data requiring for later analysis. The images are taken through HERO 4 camera and then transferred in real time to the remote processing unit such as a ground control station by the wireless connection established by a Wi-Fi router. An image processing method has been proposed to detect defects or damages such as cracks. The method based on intensity histogram algorithms to exploit the pixel group related to the crack contained in the low intensity interval. Experiments, simulation and comparisons have been conducted to evaluate the performance and validity of the proposed system.

## 1. Introduction

Inspecting structure conditions continuously of infrastructure is crucial for maintaining well operation function and serviceability for long term. As the structural material degrading is normally observed from visual structural damages, the most popular task to foresee the sign of degrading often is surface inspection. The inspection process can be achieved with three commonly approaches which using UAVs devices, ground level vehicles and directly human hands. While human hands and ground vehicles have been chosen to be the most popular use, UAVs devices has raised in interest as the ability of given highly accessibility to most of conditions, operation with flexibility and the ability to be re-programable for conducting inspection without instruction.

As the world is well developed day by day, the need of developing modern structures is crucial. One part of civil should not be underestimated is infrastructure which plays the main role in the civilisation such as economy transportation, information gathering, health service… As there is nothing against the destruction of nature and time, infrastructure is also affected by natural disasters or material degraded over decades of use. The structure requires to be investigated and maintained frequently to secure the safety for users. With structures have far distance from sea level such as bridges, investigation normally takes a lot of human effort and equipment to reveal just first scan of structure problem. Bridges inspection needs a professional inspector team, heavy lifting machines, human climbing from lethal height, resulting in multiple days inspecting time. Time taken and energy consumption of the investigation is increased with the structure size and the difficulty of the accessibility of the target area (Ariff et al. 2017) For that reason, the practice of drone, unmanned aerial vehicles (UAVs), has become more integral in investigating infrastructure. Nowadays, drone has been proven to be well perform with high agility and money saving in tasks that are dangerous for humans. UAVs has grown their popularity among military application to civil usage, such as fly camera imaging or numerous inspection quests.

Civil structures such as beam and concrete surface are always undergoing stresses, loads and natural environment that result in structures failures like cracks occurring or oxidization forming on structure surface. These failures cause the structures to reduce in strength, therefore, they may lead to structure collapse. Detection the crack in early time allow to prevent possible damage and failure. Structure crack detection is using processing methods to detect the crack within the process. There are two ways to have the crack determined which are Non-Destructive method and Destructive method to be used in the process. The evaluation of surface deficiencies condition is process with the operation of visual examination and analysis tools (Phung et al. 2017).

The measurement of the crack such as width, length also material type and number carry the early stage information to show the level of degrading of the structure could have in the future.

Automatic crack detection has been developed to have higher reliability and efficiency in surface condition detection than ordinary using the visual physical work of human. As there is no need of human hand into the practical fiend, the safety factor of the process is a big improvement of the UAVs technology. This method has pushed efficiency of Non-destructive method to higher level as manually inspecting has difficulty in accessing certain areas. These problems can be solved by using UAVs device for detecting crack which feature thermal testing device, thermal testing unit, ultrasonic device, laser device and radiographic device.

Virtual defection inspection through images for Non-destructive inspection method has been a centre of interest in civil industry nowadays. There are difficulties appearing in the testing method. These are due to the fact that there are so many shapes of crack. They are apparently not uniform and each of them show a unique shape, unpredictable size, irregular condition, shades which cause difficulty in image analysing. Because of the high potential in the industry, many virtual image analysing methods have been recommended. There are four categories of method including morphological method, integrated algorithm, percolation-based approach and practical method.

In this paper, we illustrate an inspection system for crack detection including UAVs to gather visual data of the targeting surface from infrastructure. An image processing technique is used to detect cracks from the data collected. The algorithm is based on histogram method with binary image processing. Simulations and experiments have been conducted to evaluate the validity of the proposed approach.

## 2. Literature review
In the section bellow, the recent studies in the related topics will analysed and compared to find the research gaps and thus form the aims of our project.

## 2.1 Building monitoring using UAVs
This will explore the possibility of drone in inspecting infrastructure and then examine the data collecting systems from numerous authors. The application of UAVs devices has not stopped there, drones can also be used in inspecting surfaces for testing the integrity. Baiocchi et al. (2013) have operated a quadrotor to inspect a historic structure to detect cracks and damages. The navigator component for path planner has been developed by the author to utilize the flight track, decreasing any blocking obstacles. Furthermore, 3D reconstruction of the target structure can be done by using point cloud data (Phung et al. 2016). Eschmann et al has presented a similar project although with flight control manually.

Nikolic et al. (2013) has done an inspection with UAVs on a boiler of a power plant. The project includes a system featured with trajectory-following function automatically and a custom sensor system for navigating visually. Applying all of the features, the author can inspect the inner side of the wall from the boilers, improving the navigating function of visual navigation of the drone and open denied region with GPS feature.

## 2.2 Data collection
Majority of the articles have very limited or generic information about the use of sensor in UAVs devices for inspecting infrastructure. In most cases, cameras will be used to collect the required data. Hausamann et al. (2005) has illustrated a combination of the use of infrared and optical sensors, categorize spectral bands with sensor models, a higher performance has belonged to synthetic aperture radar than colour coded cameras because of the independence regarding weather and light properties of radars. Luque-Vega et al. (2014) have processed the project with cameras which have thermal and infrared feature for increasing the effectiveness of object detection by background removing. Larrauri et al. (2013) has utilized HD camera for their experiment to process 4 fps data. Eschmann et al has done a project about navigating using GPS signal together with a commercial camera which has the quality of 12-PM for post-processing offline. Nikolic et al has considered about a CMOS virtual sensor and economic sensor with and personalize-design circuit which helps UAV to collect data and navigate in environments that lost signal of GPS. From results collected from these above projects, the colour coded cameras are primarily used for achieving data. The performance of colour camera and the cost of it are two opposite values, finding the right ratio between those two can lead to achieving great results without sacrificing too much budget. With all of the circumstances discussed above, camera with infrared feature can be additionally applied which can increase the effectiveness of detection.

Whitehead et al has analyse different sensor models applied to remote sensor use. From the idea of vision solution, shows a variety of benchmarks and datasets for optical flow methods, and talks about edge detection technique in detail. About controlling method, the author has shown the method for low-level control and stabilization, flying from regular linear pathway to more complicated nonlinear controller with high accuracy, while mentions about high level control. There are other opinions have considered about the planning method under the view of uncertainties. The book about planning in robotic also shows the constraints of low-level dynamics on planning.

## 2.3 Crack detection using vision-based methods

With detection using UAVs techniques, processing collected images is an essential requirement to analyse defects or cracks. Regarding image with large datasets processed from the system, automatic machine learning and artificial intelligence algorithms are the methods usually been applied to analyse sections of dataset to utilize in training purpose and solve the data remained (Phung et al., 2019; Chen et al., 2017; Phung et al., 2017; Shi et al., 2016; Amhaz et al., 2016; Oliveira and Correia, 2013). The method usually solves existing datasets without any problem although may have a chance to struggle with arbitrary. About algorithms for crack segmentation, the essential requirement of the method is generality, e.g., to examine the changing colours and shapes of the target objects. So far, there are numerous algorithms about segmentation on detecting the change of crack shapes have been achieved. The method that usually been use in general for image segmentation is the binarization algorithm shown in (Otsu et al. 1979) that processed through the intensive histogram of the image to analyse the optimised threshold. With the visual impact, the relocation of image continuous intensity with histogram and smoothing techniques can be performed to be an enhancement method (Kwok et al. 2011, Dinh et al. 2019), selecting the correct threshold holds an important role in increasing the crack detection accuracy using images. Lately, the threshold is refined from the old version of Otsu's algorithm to the new improvement version interactive tri-class thresholding technique (ITTT) from Cai et al. (2014). The ITTT method will firstly segment the picture into background and object using an initial threshold regarding the picture histogram. Then the darker half from the background is merged to form a new region together with the lighter half from the object. The process will be repeated continuously until the threshold of current and previous experiment are lower than a pre-assumed number. Both ITTT and Otsu algorithms work effectively on birarizing pictures although they show low effectiveness of analysing low intensity images about surface inspection.

## 3. Methodology

The following part will discuss about the procedure of the data collecting process as well as how images collected from field experiments are processed using the proposed algorithms.

### 3.1 Data collection Using UAVs

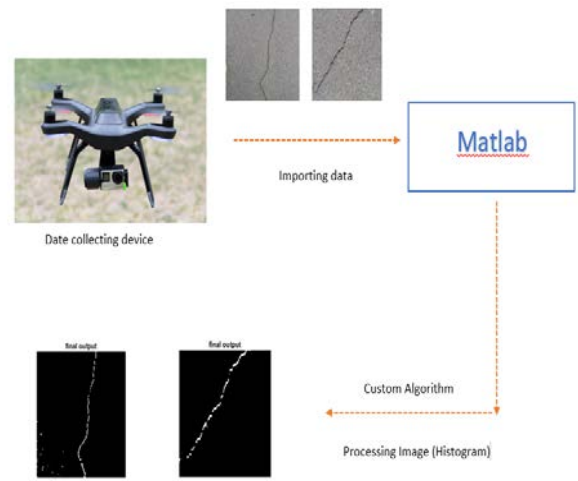

Figure 1: Collecting Data Controlling Map

Figure 1 illustrates the system which includes a UAV (3DR Solo model) to collect images from the targeting surface. The visual data will be obtained by a camera installed on the drone. As the drone flies to the target location, the camera will be controlled manually from the user on the ground to take pictures of the targeting areas. The collected images are then processed with a histogram-based algorithm to remove the background and noises from the orginal images to identify cracks.

### 3.1.1 Drone 3DR Solo

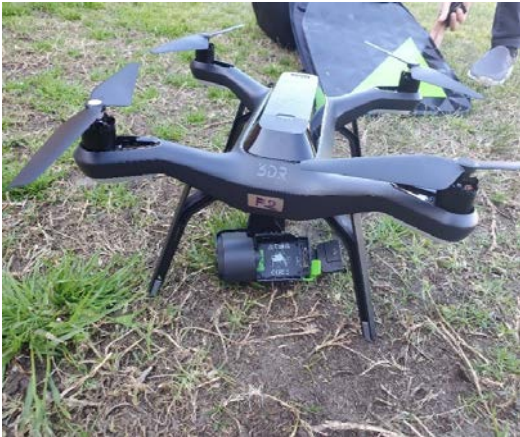

Figure 2: 3DR Solo Drone with Gimbal Bay Attached

The drone used in Figure 2 is a 3DR Solo featured colour camera and manual controlling devices. The 3DR Solo drone also has an open source for programmer (SKDs and APIs) which can offer adjustments in program to suit any purposes that require drone. The design of Solo drone is comparable with GoPro camera by inserting the camera to the gimbal bay appeared at the bottom of the drone. The manufacturer has featured Solo model with the best automated camera shots (Smart Shots) on the market. The Solo plays as a remote data collecting device which can flies to an unreachable high for human without any assisting equipment.

### 3.1.2 Ground Control Station and Controller
The ground control station shown in Figure 3 will be set up to perform the data collection from the drone. The ground station helps to the users to communicate with the UAVs and to give orders to the drone.

*A. Remote Controller*

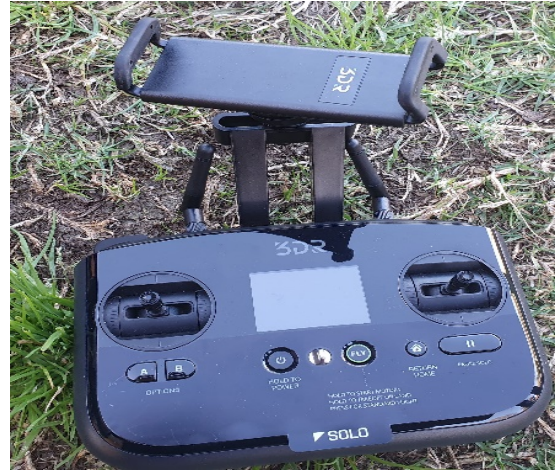

Figure 3: Controller handle

The remote controller shown in figure 3 is compatible with mobile phone to provide a complete visual controller. The controller is designed with a mobile holder to maximize the ability of controlling and monitoring. The controller has a simple design with optimize direction of the drone and the movement of the gimbal to control the collecting data procedure.

*B. ArduPilot Open Source Software*
The drone comes with software which can be installed to a computer platform to control and receive the data collected from the accessory featured on the drone and information achieved from GPS signal for analysis. The software is called ArduPilot as shown in Figure 4 which is an open source navigation and autopilot software. ArduPilot is an open source autopilot software which is available with the most developed full-featured and high reliability.

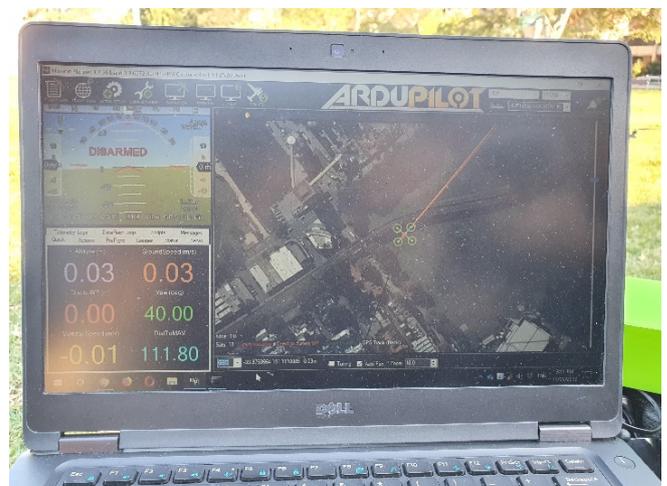

Figure 4: ArduPilot Software Operating on Laptop

ArduPilot is the only open source autopilot software that is able to control any remove vehicles visually from submarines, helicopters, multirotor to conventional airplanes. The designing team even improve the software further to support the controlling of new developed vehicles such as compound helicopters and quadrotors. The software has been estimated to be installed over 1 million devices over the world and with the advantage of simulation and analysis tools, data directly import-export, the software is the most popular in use for autopilot purpose software. The software can be found in many OEM UAV manufacturer including iDrones, Kespry, AgEagle, PrecisionHawk and 3DR. It is also tested and used for development purpose by famous corporations and institutions including Insity/Boeing, Intel and NASA.

### 3.1.3 GoPro Colour Camera

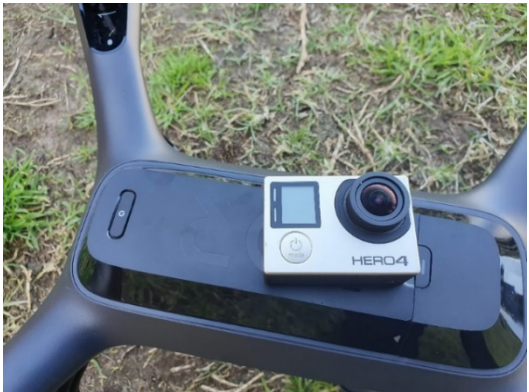

Figure 5: GoPro HERO4 Camera

The HERO4 from GoPro in Figure 5 is used for collecting visual data. It is featured with a f/2.8 focus free and fixed maximum aperture and 12 megapixel CMOS with added Bluetooth connect availability, Protune Available, Highlight tag for new processor and photo when compare to lower model HERO3+, double frame rating in resolutions in most of cases. The camera can record at the 24, 25 and 30 fps frame rate with 4K UHD video quality (3840x2160). The camera also offers a Superview mode that can record at the frame rate of 25 fpt with 4K quality. Moreover, it also has many different resolutions and frame rate available for flexibility. The camera used in this project can shoot stably at maximum quality of 12 megapixel with frame rate of 30 fps. The battery of the camera can stay up to 65 minutes when the camera is disabled Wi-Fi when shooting at 4K quality on 30 fps and can be increased to reach 110 minutes at 720 MP at 240 fps.

### 3.1.4 Camera Gimbal Bay

The gimbal bay as shown in Figure 6 is mounted to the UAV to provide shock absorbing ability for taking photos or filming, avoid shaking or vibrating. The gimbal is featured with three motors to support 3-axis rotating. By featuring a gimbal bay, the camera can be operated in all axes as the rotor of the bay is moving. The stabilized ability of the bay is guided by algorithms, it can define the differences between shaking and tracking shots movements. This helps the camera to work with ease on the air regardless acceptable wind conditions.

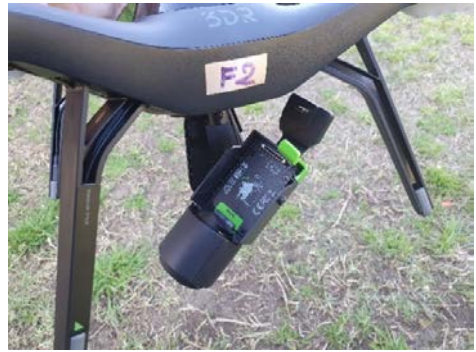

Figure 6: Gimbal Bay Attached on The Drone

### 3.2 Crack detection algorithm

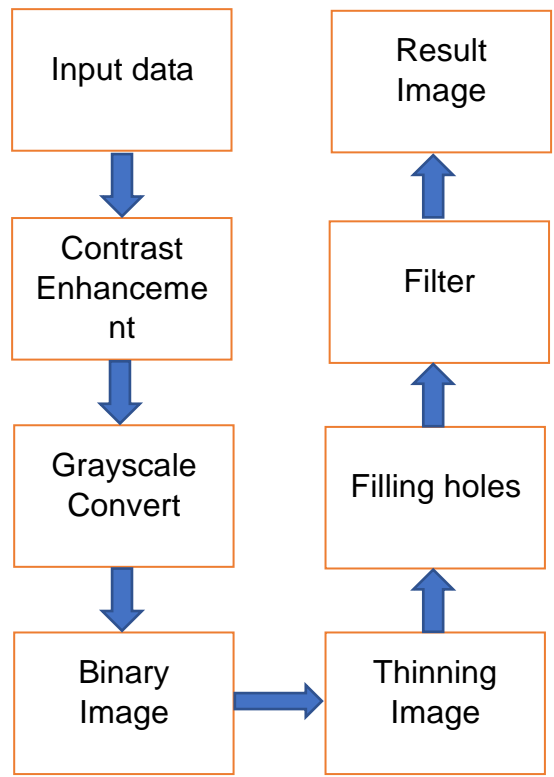

Figure 7: Diagram of the crack detection algorithm

Figure 7 shows a general view of the algorithm used in the project. The visual data collected from the field poses challenges for identifying cracks. An algorithm has been developed to process images to define crack for increasing efficiency for detecting crack with involved fewer human hands. The algorithm will be developed using histogram information.

The crack from the visual data consumed only a small number of pixels when compare to the background. The crack regions will be assumed to have a darker colour when compare to the background and the thresholding algorithm should be focusing on the dark areas.

### 3.2.1 Contrast stretching

The first step used in the algorithm is contrast enhancement. This step considers the gray colour value (scalar) of the image *f* including value *f(x)* that has the possibility to be a subset of the gray value regarding the range of visual data.

The result is the grayscale image that utilizes all shade of gray ranging from 0 to 255:

$$g = \frac{f - f_{min}}{f_{max} - f_{min}} \qquad (1)$$

where $f_{min}$ and $f_{max}$ are respectively the minimum and maximum gray values of the input image while *f* is the input gray value.

By assuming the crack to have darker colour, the enhancement of contrast will make the differences between the dark and light colour regions in the picture further. This will be a preparation for the developing of the algorithm later when trying to isolate the crack. The algorithm of contrast stretching will organize value of intensity of the image in grayscale. The technique will saturate the top 1% and the bottom 1% of the pixel values based on the intensity of the pixel. This will result in the improving about the contrast of the picture. The method will be finished with the crack region appearing darker (more obvious for computer recognition) and the background will be lighter therefore, the background can be easily removed by applying thresholding technique.

### 3.2.2 Converting RGB to grayscale

By adapting the picture in grayscale to be the second step, the data will be change from the basic colours which are red, blue and green into difference shades of gray. What the technique does is converting the real colour of the image RGB into grayscale by removing the saturation and the hue information while remain the luminance.

Weighted method which is usually used for converting grayscale picture has shown its advantage of better illustration in grayscale image than other relative methods. The red colour has longer wavelength when compares with other two colours. The green colour beside of having lower wavelength compare to the red colour, it also has the virtually soothing effect. By the information achieved, the contribution of colours has been adjusted as equation below

The weighted method can be described:

$$f(x,y,z) = 0.3x + 0.59y + 0.11z \qquad (2)$$

Where f(x,y,z) is the luminance function with 3 variance x,y,z respectively represent the value of red, green and blue colour from the image.

The equation above shows that the contribution of three colour which are Red, Green and Blue have taken the ratio of 30%, 59% and 11% respectively regarding their effect properties mentioned above.

The purpose of converting the input picture into grayscale is to create a sub-step before converting to binary image for further isolating the crack from the background pixels.

### 3.2.3 Binary image converting (thresholding)

Thirdly, binary image is defined as a digital visual data (image) that can obtain only two possibilities of value for a pixel unit. Black and white are the two typical colours used in binary image. The method of colour selection in binary image process will be divided into foreground colour as object and the remaining colour to be background colour. Thresholding will play the mail role in revealing the crack by isolating the crack pixels from the background.

The primary goal of image segmentation procedure is to group visual data into subsets (also called regions, classes) that are homogeneous regarding their features or properties. There are numerous of image segmentation methods have been recognized to optimize processing steps for digital image. The chosen technique for this project for region segmentation method is thresholding. Thresholding technique will select a threshold and group in to two group: group contained pixel with value less than the chosen threshold and the second group with higher or equal value to the threshold. The equation below will illustrate how thresholding working.

$$g(x,y) = \begin{cases} 0 & f(x,y) < f_c \\ 1 & f(x,y) > f_c \end{cases} \qquad (3)$$

where fc is the threshold value, g(x,y) is the output intensity and f(x,y) is the intensity at (x,y).

### 3.2.4 Images thinning
The thinning algorithm will delete pixels to make the targeting without holes object shrinks to minor connected line and for the option with holes object, the morphological process will shrink the object to linked circle halfway between the outer boundary and each hole. This method will save the Euler number. The thinning morphological processing on a visual data f relative to structuring element s is written in the following mathematic notation

$$f \otimes s = f \setminus (f \odot s) \qquad (4)$$

$f \otimes s$ is the morphological thinning command with a structure of element s within image f. The note \ shows the difference between difference unit in group. The hit-or-miss transformation is represented by symbol $\odot$.

From the definition of the structuring elements in the previous part, the thinning process will be applied as:

$$d_0 = f \qquad (5)$$
$$d_k = d_{k-1} \otimes \{s_1, \cdots, s_8\} \qquad (6)$$
$$= (\cdots ((d_{k-1} \otimes s_1) \otimes s_2) \cdots) \otimes s_8 \qquad (7)$$

The K value will be iterated until $d_K = d_{K-1}$ is reached and $d_K$ will be the result for thinning.

### 3.2.5 Filling holes
The crack object after thinning will reduce in overall size to become thin line. The process mostly results in a line object to determine the crack position however, when the crack area has significant large size holes inside the damage structure which may lead to imperfection crack image or result to error in identifying cracks. This imperfection usually is called holes which is often observe after segmentation process. The solution to solve this problem has been propose as holes filling function to fill the crack miscounting from thinning step. The hole filling algorithm will perform a morphological process within the input image. The filling holes function will operate an interior filling process which will change all value of a pixel to 1 if surrounding pixel are 1.

### 3.2.6 Filter
The median filter is applied to smoothing the signal of images with nonlinear filter. The filter is excellent for application such as impulsive type noise removal from signals without removing the edge of included of the objects in the image. There are numerous of this filter variations, and the most popular for removing speckles and noise from images is the two-dimensional median filter. The two-dimension median filter is performing image filtering in two dimensions to result the output of the median value of the size 3 by 3 to the related pixel surrounding the targeting pixel in the examining image. The nonlinear function equation of the median filter can be proposed as:

$$Y(n) = med\ [x(n - k),\ x(n-k+1),\ldots,\ x(n),\ldots,x(n + k - 1),\ x(n + k)] \qquad (8)$$

With y(n) and x(n) are the output and input signal respectively, the median filter gathers an area including N=2k+1 pixel of the input date and the executes the filter function on the group of data. The filter length N=2k+1 is the only one design parameter in the function, otherwise, the median filter function is in simply standard form.

## 4. Result and discussion
The results from this study are now presented and followed with a discussion on each related target.

### 4.1 Experimental setup

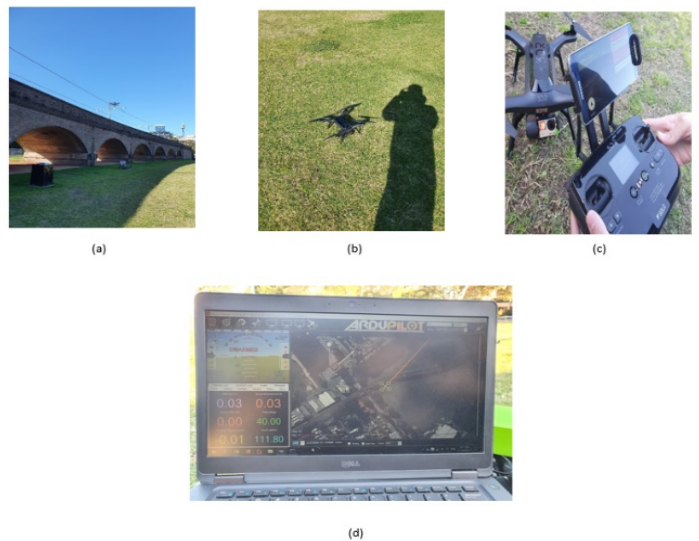

Figure 8: (a) targeting structure; (b) 3DR Solo drone on the testing field; (c) controller device

with phone attached as a visual data receiver; (d) ground control station overview.

The drone in the experiment flies to the targeting structure which is the base structure of light rail system that built across Wentworth park located in Sydney Central as illustrated in Figure 8. The light rail system bases at Wentworth park has been built since 1997 so there is a high chance to find structural crack in the system. The location also convenient as there is no traffic in the testing zone therefore, the visual control station can be located right under the targeting structure. The testing time was under low wind condition and appropriate light condition. The testing field was chosen based on the historic infrastructure that may have damages occurred and low to no high attitude blockage. The drone was controlled manually to take images of the areas that have high possibility of crack occurring. The controller was connected to mobile phone during the process as a visual controlling method. The possibility of crack was determined by observation by the user through visual control station using mobile phone.

## 4.2 Inspection system and approaching method

This section will discuss about the set up between data collecting devices and related accessories.

### 4.2.1 Gimbal bay performance

The inspecting system has shown satisfaction during the field-testing period. For collecting visual data, the gimbal bay has proved the ability of stabilizing during the flying time. The images collected from the setup of gimbal bay and GOPRO HERO4 have showed acceptable quality for further analyse with picture processing stage.

The pictures achieved are in excellent quality. There is no blurring appeared in the images during the data collection process. Gimbal bay also proves to have a strong holding ability to the camera attached with it. The hook design also shows its sufficient holding ability to keep the camera device locked in the bay. The primary material used in gimbal bay is plastic therefore it can withstand high moisture environment, high resistance to material corrosion. The material chosen shows that the design has consider about weight factor for flying device therefore, with plastic or ABS specifically, it has enabled energy saving and durability features in the product.

However, the design of the gimbal bay still has gap that can let moisture to access to the electrical part of the camera and the bay itself. If the design can be reconsidered to have gaskets or solutions to block moisture out of the connection, the accessory will be more favourable in the market for UAVs.

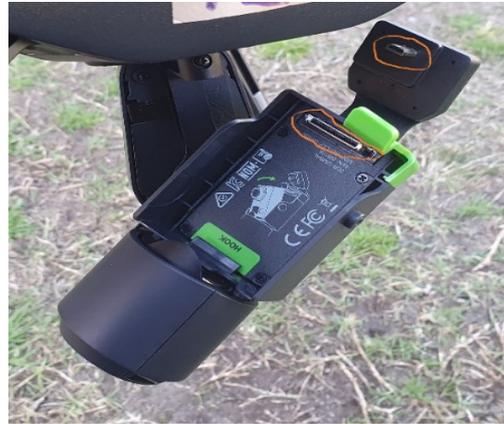

Figure 9: Illustration of Exposing Areas

### 4.2.2 Drone Performance

The drone also successfully delivers the camera into the targeting position under mild weather conditions. The drone has four motors located at 4 corners of the device therefore, the drone is able to maintain the stability up to a certain level under unfavourable weather conditions. The four motors of the drone have generated with an identical level of noise with is quite common in operating UAVs devices. The level of noise is acceptable under subjective evaluation according to an opened testing environment. The factor of noise needs to be reconsidered when the UAVs device is operated under high population area which can leads to uncomfortable reactions from surrounding citizen. The factor also needs to be critically aware when flying at night-time and early in the morning as the previous mentioned factor.

The motor housing design is simplified to save unneeded material to optimize weight saving for the drone. The outlook of the housing design shows that the manufacture has managed to increase the aesthetic feature to the perfection. The housing has consistent gap between the connections of different parts of the shell. The condition above prove the manufacture factory has necessary technology to minimize the clearance or error of processing machine.

Although the design of the motor housing is sufficient for operating the drone, there is still possibility to improve the design for better performance and its popularity among users of UAVs devices. The housing has revealed an expose of the wire core of the motor which has let a big area for moisture and obstacle to be trapped inside and cause damaging to the motor. If the design can be further improved with this area fully covered and moisture blocked, the life span of the device will be significant extended.

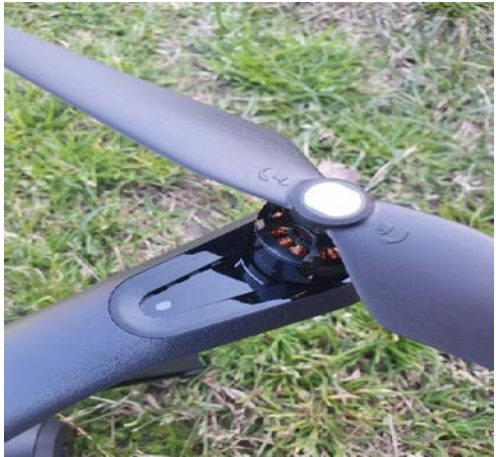
Figure 10: Rotor Housing

### 4.2.3 Controller devices
The controller acts as middle step between giving commands to the drone and collecting data transferred from the camera to the storing devices (computer or mobile phone). The design of the controller also received intention from the engineers. The design of the controller has very high glossing surface showing great artistic look to the users. All the buttons are responsive during operation and the function of each button is carefully defined from the manufacturer to make sure they can sufficiently and efficiently support the users. The phone holder frame has a been created with minimalism thought to show the wisdom in material use.

The controller has a square shape which has a low rating for holding experience. The shape is observed to have very low rating regarding the contact between the device and user's hands. the square shape of the housing has heavily consumed material for unnecessary area. The polish surface and shape are slippery or unable to hold with ease when there is moisture involved. The idea of improvement for the controller in opinion can be observed from gaming handle. The ideal handle having a better ergonomic grip compare to the testing handle. The handle should have a surface treatment with a thin layer of rubber to increase grip and comfortability of the users.

### 4.3 Crack detection result
Seven images taken from the drone are used as inputs to the proposed algorithm to detect cracks. The results are as follows.

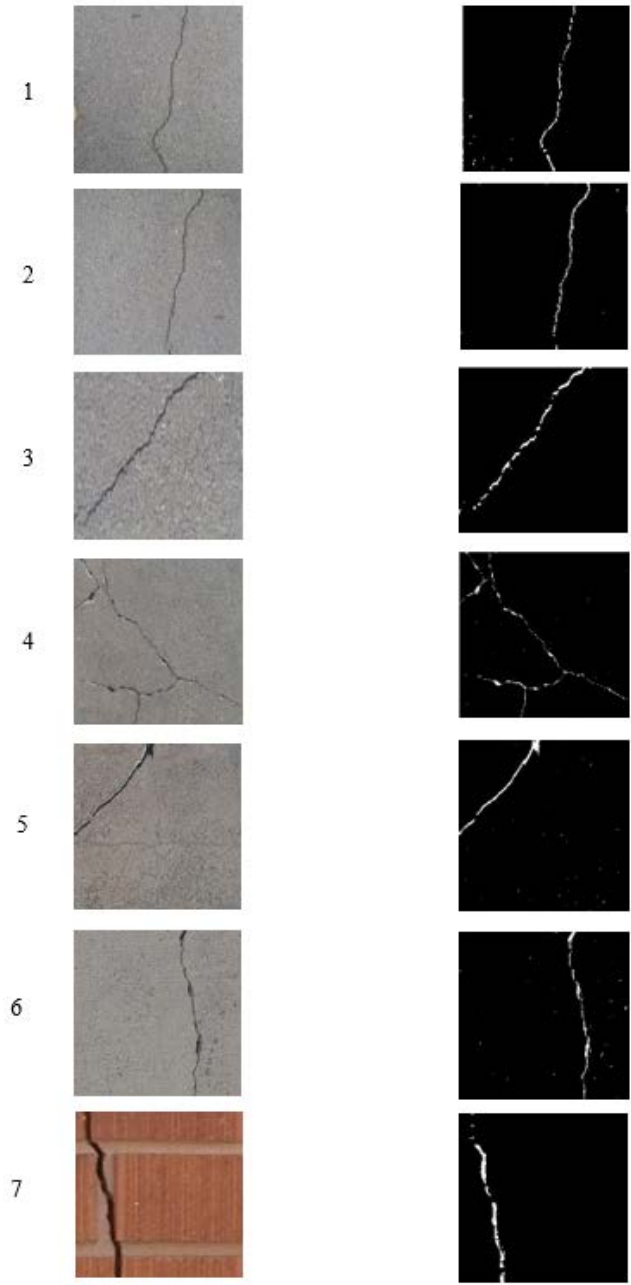
Figure 11: Image Processing Results. First column: original image; Second column: result image.

The images resulted from the image processing algorithm have clearly or partially defined the cracks from the surface of the targeting structure. By observation, image 5 shows the best performance among the results with the crack fully defined by the algorithm. The detail of image 5 matches the original input image when put several features of the picture under consideration to determine the performance of the image processing method. The thickness of the crack is corresponded to the input image which proves the efficiency of the thinning and filling algorithms regarding the features in the original image. Although there is appearance of noises, the problem is insignificant when considering the total pixels of the image and the crack compared to the noise.

The group of pictures 1, 2, 3, 4 and 6 also reveal most of the characteristics of the crack included in the image. However, the proposed algorithm has failed to carry all of the dimensions of the damage out to the paper. The cracks from the result pictures have several disconnected pixels which can lead to the misleading of identification of the cracks. The problem can further result to underestimation the significance of the damages and this will eventually lead to a fail structural investigation. Image 1 and 6 shows a much more noises when compare to image 2, 3, 4. The problem maybe a result of binary transformation process. There is a high possibility that the background contains a high amount of dark area which is linked together. This cause error to identification process. The problem reveals a point for further improvement of the algorithm.

The image 7 was taken from a brick textured surface to test the flexibility of the algorithm in solving different backgrounds. The result image shows an impressive illustration of the crack in the structure that clearly stands out from the background. The result proves that the chosen image processing method can process with some certain background with textiles such as brick which is one of the most commonly seen in infrastructure. There are still some unsolved problems appeared in the result. The dimension of the crack has not been fully discovered after running with the processing algorithm. There is missing in thickness and connectivity from the result which need to be considered in the development of the algorithm in the future.

The proposed algorithm has showed an appropriate performance when compare to expecting result in visually defining damages from structure surface with image data. Cracks have been revealed with or without their full dimension, but the identification of cracks can still be identified with observation as the level of standing out compare to the background. Unfortunately, the algorithm cannot remove all of the noises that cause from error in converting from coloured images to binary images. The problem can be solved in the future research by propose a more suitable algorithm for image filtering or cleaning to ensure the background pixels are fully identified by the algorithm. The connectivity of the crack is also aa error cause by miscounting of object pixels. Further development in pixel grouping algorithm may show a high chance in solving the error.

## 5. Conclusion

In this preliminary investigation project, a method of infrastructure investigation procedure using drone has been proposed to optimise the work with remote controlling method. There are two stages included in the investigation procedure as data collecting from targeting infrastructure using drone (UAVs devices) and data processing using custom algorithm developed from histogram method using MatLab.

Data collection was done by 3DR Solo drone and GOPRO HERO4 camera with gimbal bay stabilizing support under the light rail basement located at Wentworth park. The testing day was chosen under consideration of weather factors such as rain and wind conditions. The clear images from the targeting structure were obtained under low wind and no rain conditions. An algorithm has been proposed to act as image processing technique base on histogram background. The results have been received after the import the input images through image processing through MatLab with several image adjustment steps. The preliminary results are able to define the natural form of the cracks although there is missing of dimension of the crack and several images still contain noises.

From the preliminary result, the method proposed can be improved to achieve clearer images (noise removal), better crack dimension in length and thickness. There are also several options to further developing the inspecting method. Automatically crack detection applying artificial intelligence and image processing are of great considerations for our future work.

## Reference

Efford, N. (2000). *Digital Image Processing: A Practical Introduction Using Java$^{TM}$* , Pearson Education

Drone controller from PARROT ANAFI Extended Drone with Controller, viewed 18 October 2019, <https://brain-images-ssl.cdn.dixons.com/6/1/10187816/l_10187816_005.jpg>

Oliveira, H. & Correia, P. L. (2013), *Automatic Road Crack Detection and Characterization*.

IEEE Transactions on Intelligent Transportation Systems, pp. 155–167.

Phung, M.D., Quach, C.H., Tran, H.D., and Ha, Q. (2017), *Enhanced Discrete Particle Swarm Optimization Path Planning for UAV Vision-based Surface Inspection*, Automation in Construction, vol.81, pp. 25–33.

Amhaz, R., Chambon, S., & Baltazart, V. 2016, *Automatic Crack Detection on Two-Dimensional Pavement Images: An Algorithm Based on Minimal Path Selection*, IEEE Transactions on Intelligent Transportation Systems, pp. 2718–2729.

Shi, Y., Cui, L. M., Qi, Z. Q., Meng, F., & Chen, Z. S. 2016, *Automatic Road Crack Detection Using Random Structured Forests*, IEEE Transactions on Intelligent Transportation Systems, pp. 3434–3445.

Chen, J. H., Su, M. C., Cao, R. J., Hsu, S. C., & Lu, J. C. 2017, *A self-organizing map optimization based image recognition and processing model for bridge crack inspection*, Automation in Construction, pp. 58–66.

Hoang, V. T., Phung, M. D., Dinh, T. H. and Ha, Q. P. (2019), *System Architecture for Real-Time Surface Inspection Using Multiple UAVs*, IEEE Systems Journal, doi: 10.1109/JSYST.2019.2922290.

Otsu, N. 1979, *A Threshold Selection Method from Gray-Level Histograms*, IEEE Transactions on System, pp. 62–66.

Kwok, N. M., Jia, X. P., Wang, D., Chen, S. Y., Fang, G., & Ha, Q. P. 2011, *Visual Impact Enhancement via Image Histogram Smoothing and Continuous Intensity Relocation*. Computers and Electrical Engineering, pp. 681–694.

Cai, H. M., Yang, Z., Cao, X. H., Xia, W. M., & Xu, X. Y. 2014, *A New Iterative Triclass Thresholding Technique in Image Segmentation*. IEEE Transactions on Image Processing, pp. 1038–1046.

Phung, M. D., Quach, C. H., Chu, D. T., Nguyen, N. Q., Dinh, T. H. and Ha, Q. P. (2016), *Automatic interpretation of unordered point cloud data for UAV navigation in construction*, *2016 14th International Conference on Control, Automation, Robotics and Vision (ICARCV)*, Phuket, pp. 1-6, doi: 10.1109/ICARCV.2016.7838683.

Li, H., Wang, B., Liu, L., Tian, G., Zheng, T., & Zhang, J. 2013, *The design and application of SmartCopter: An unmanned helicopter-based robot for transmission line inspection*. Proceedings of the IEEE Chinese Automation Congress (CAC), Changsha, China, pp. 697–702.

Zhang, J., Liu, L., Wang, B., Chen, X., Wang, Q., & Zheng, T. 2012, *High speed automatic power line detection and tracking for a UAV-based inspection*, Proceedings of the IEEE International Conference on Industrial Control and Miscs Engineering (ICICEE), Xi'an, China, pp. 266–269.

Luque-Vega, L.F., Castillo-Toledo, B., Loukianov, A., & Gonzalez-Jimenez, L.E. 2014, *Power line inspection via an unmanned aerial system based on the quadrotor helicopter*, Proceedings of the IEEE 17th Mediterranean Electrotechnical Conference (MELECON), Beirut, Lebanon, pp. 393–397.

Larrauri, J., Sorrosal, G., & González, M. 2013, *Automatic system for overhead power line inspection using an Unmanned Aerial Vehicle - RELIFO project*, Proceedings of the IEEE International Conference on Unmanned Aircraft Systems (ICUAS), Atlanta, GA, USA, pp. 244–252.

Martinez, C., Sampedro, C., Chauhan, A., & Campoy, P. 2014, *Towards autonomous detection and tracking of electric towers for aerial power line inspection,* Proceedings of the IEEE International Conference on Unmanned Aircraft Systems (ICUAS), Orlando, FL, USA, pp. 284–295.

Baiocchi, V., Dominici, D., & Mormile, M. 2013, *UAV application in post-seismic environment. Int. Arch. Photogramm. Remote Sens. Spatial Inf. Sci*, pp. 21–25.

Eschmann, C., Kuo, C., Kuo, C., & Boller, C. 2012, *Unmanned aircraft systems for remote building inspection and monitoring.* Proceedings of the Sixth European Workshop on Structural Health Monitoring, Dresden, Germany.

Nikolic, J., Burri, M., Rehder, J., Leutenegger, S., Huerzeler, C., & Siegwart, R. 2013, *A UAV system for inspection of industrial facilities*, Proceedings of the IEEE Aerospace Conference, Big Sky, MT, USA, pp. 1–8.

Hausamann, D., Zirnig, W., Schreier, G., & Strobl, P. 2005, *Monitoring of gas pipelines—A*

*civil UAV application*, Aircraft Eng, Aerosp. Technol, pp. 352–360.

Whitehead, K., & Hugenholtz, C.H. 2014, *Remote sensing of the environment with small unmanned aircraft systems (UASs)*, part 1: A review of progress and challenges 1. J. Unmanned Veh, Syst. pp. 69–85.

Shi, J., & Tomasi, C. 1994, *Good features to track*, Proceedings of the IEEE Computer Society Conference on Computer Vision and Pattern Recognition (CVPR'94), Seattle, WA, USA, pp. 593–600.

Dinh, T. H., Phung, M. D. and Ha, Q. P. (2019), *Summit Navigator: A Novel Approach for Local Maxima Extraction*, in *IEEE Transactions on Image Processing*, vol. 29, pp. 551-564, doi: 10.1109/TIP.2019.2932501.

Bay, H., Tuytelaars, T., & van Gool, L. 2006, *Surf: Speeded up robust features*, Computer Vision—ECCV 2006; Springer: Berlin/Heidelberg, Germany, 2006; Volume 3951, pp. 404–417.

Ariff, O.K., Marshall, D.M., Barnhart, R.K., Hottman, S.B., Shappee, E., & Thomas Most, M. 2017, *Introduction to Unmanned Aircraft Systems,* Boca Raton, FL, USA.

Phung, M.D., Hoang, V.T., Dinh, T.H., and Ha, Q. (2017), *Automatic crack detection in built infrastructure using unmanned aerial vehicles*, 34th International Symposium on Automation and Robotics in Construction (ISARC), pp. 823-829, Taipei, Taiwan.